\documentclass[10pt,twocolumn,letterpaper]{article}

\usepackage{iccv}
\usepackage{times}
\usepackage{epsfig}
\usepackage{graphicx}
\usepackage{amsmath}
\usepackage{amssymb}
\usepackage{outlines}
\usepackage{caption}
\usepackage{subcaption}
\usepackage{float}
\usepackage{booktabs,arydshln}   
\usepackage{adjustbox}
\usepackage{flushend}

\usepackage[pagebackref=true,breaklinks=true,letterpaper=true,colorlinks,bookmarks=false]{hyperref}

\iccvfinalcopy % *** Uncomment this line for the final submission

\def\iccvPaperID{5409} % *** Enter the ICCV Paper ID here

\ificcvfinal\fi

\newcommand\minisection[1]{\vspace{1mm}\noindent \textbf{#1}}

\newcommand{\data}{{OAK}\xspace}
\newcommand{\datafull}{{Objects Around Krishna}\xspace}
\newcommand{\kcam}{{KrishnaCam}\xspace}

\begin{document}

%%%%%%%%% TITLE
\title{Wanderlust: Online Continual Object Detection in the Real World}

\author{
    Jianren Wang$^1$~~~~
    Xin Wang$^2$~~~~
    Yue Shang-Guan$^{3}$~~~~
    Abhinav Gupta$^1$
    \\\\
    $^1$\normalsize Carnegie Mellon University~~~~$^2$\normalsize Microsoft Research~~~~
    $^3$\normalsize University of Texas, Austin\\
}

% \maketitle
% % Remove page # from the first page of camera-ready.

\twocolumn[{
\maketitle
\begin{center}
\includegraphics[width=\textwidth]{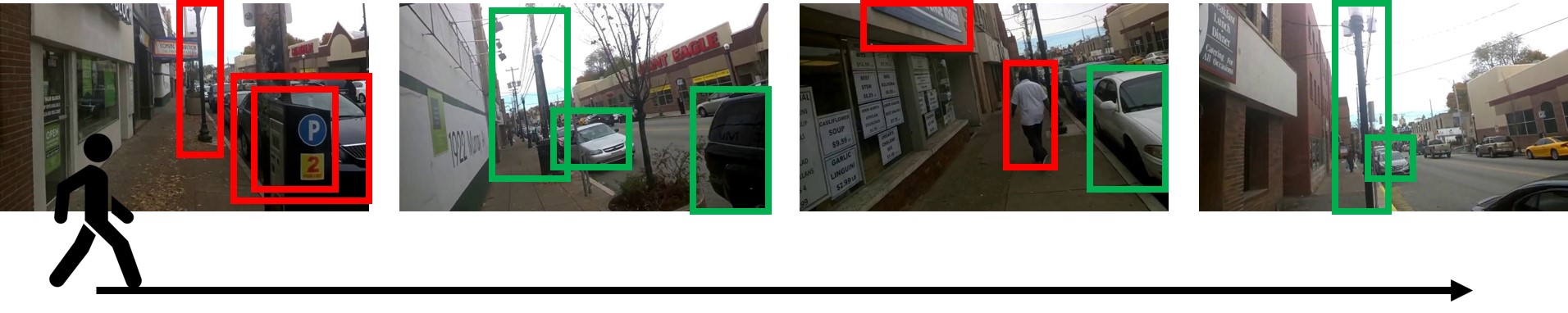}\\
\captionof{figure}{\textbf{Wanderlust}: Imagine an embodied agent is walking on the street.  It may observe new classes and old classes simultaneously. The agent needs to learn fast given only a few samples (red) and recognize the subsequent instances of the class once a label has been provided (green). In this work, we introduce a new online continual object detection benchmark through the eyes of a graduate student to continuously learn emerging tasks in changing environments. 
}
\label{fig:teaser}
\end{center}
}]

%%%%%%%%% ABSTRACT
\begin{abstract}
% Researchers have made significant advances in the field of object detection. However, most of these advances are trained and evaluated on static datasets in an offline setup. We anticipate that 
Online continual learning from data streams in dynamic
environments is a critical direction in the computer vision field. However, realistic benchmarks and fundamental studies in this line are still missing. To bridge the gap, we present a new online continual object detection benchmark with an egocentric video dataset, Objects Around Krishna (OAK).  OAK adopts the KrishnaCAM videos, an ego-centric video
stream collected over nine months by a graduate student. OAK provides exhaustive bounding box annotations of 80 video snippets
($\sim$17.5 hours) for 105 object categories in outdoor scenes. The emergence of new object categories in our benchmark follows a
pattern similar to what a single person might see in their day-to-day life. The dataset also captures the natural distribution shifts as the person travels to different places. These egocentric long running videos provide a realistic playground for continual learning algorithms, especially in online embodied settings. We also introduce new evaluation metrics to evaluate the model performance and catastrophic forgetting and provide baseline studies for online continual object detection. We believe this benchmark will pose new exciting challenges for learning from non-stationary data in continual learning. The OAK dataset and the associated benchmark are released at \url{https://oakdata.github.io/}.
\end{abstract} 

%%%%%%%%% BODY TEXT
\section{Introduction}
Modern object detectors have made substantial progress on internet images~\cite{carion2020end,he2017mask, ren2015faster}. Nevertheless, challenges remain when detecting small
objects~\cite{cheng2016survey}, scaling to a large number of categories~\cite{gupta2019lvis} or learning from only a few labeled examples~\cite{kang2019few,wang2020frustratingly}. The detector often degenerates significantly when deployed on robots or ego-centric videos in an embodied environment~\cite{damen2018scaling}.   

If we take a closer look at the typical learning setup, most of the advances in object detection have been realized using static images in an offline learning setup. In this setup, the data is labeled with a fixed set of categories and divided into two parts: training and testing. There is a training phase where the detectors are learned by randomly shuffling and feeding the training data for hundreds of epochs, followed by an evaluation on the test set. However, this offline training and evaluation setup often does not reflect how humans or embodied AI agents learn. 

 Unlike the current \emph{static offline} settings, humans receive a continuous temporal stream of visual data and train and test the model on the same visual data, which is an \emph{online continual} setting.  The categories of interest are unknown beforehand. The model needs to learn new object categories when objects belonging to previously unseen categories appear. Most of the learning happens online; we cannot use the training data repeatedly across hundreds of epochs.  
 
 A side effect of this online continual learning setting is catastrophic forgetting~\cite{mccloskey1989catastrophic}. Though previous works~\cite{aljundi2018memory, kirkpatrick2017overcoming, lopez2017gradient, schwarz2018progress, zenke2017continual} attempt to address the issue, they are usually evaluated offline and do not work well on structural prediction tasks like object detection. What hinders the progress in online continual learning is the lack of realistic datasets and benchmarks. Most of the current research~\cite{perez2020incremental, shmelkov2017incremental, zhou} re-purpose existing static datasets such as VOC and COCO to evaluate continual object detection. These approaches use object categories one by one in a sequential manner. These manual splits and artificial setups differ from the scenarios often encountered by embodied agents, where the emergence of new tasks often follows the trajectories of the agents and the frequencies of the object instances vary from task to task.  For example, the agents might observe the instances of the same category after a few hours or even days. They may visit some objects more often than others and revisit previously observed objects.

In this paper, we present a new online continual object detection benchmark. Our benchmark consists of a new labeled dataset  -- \textbf{\data} (Objects Around Krishna). \data uses the videos from the KrishnaCam~\cite{Singh16Krishn} dataset -- an ego-centric video dataset collected over nine months of a graduate student's life. \data contains 80 labeled video snippets totaling around 17.5 hours (roughly 1/4 of the raw videos in KrishnaCam) with bounding box annotations of 105 object categories in outdoor scenes. \data provides a natural data distribution and the task emergence following the trajectories of a single person. A few objects frequently appear due to redundancy in daily routines, while new objects and categories constantly appear as they visit various places. This dataset is a realistic playground to study online continual learning, enables the embodied agent to learn from a human's experience, and provides a unique opportunity for researchers to pursue the essence of lifelong learning by observing the same person in a long time span. 

We introduce several new evaluation metrics in the online continual learning setup. In contrast to the previous task incremental or class incremental settings in continual learning, there is no explicit task boundary in our setup and new tasks emerge following the temporal order in the videos. Therefore, we evaluate the overall performance (continual average precision, CAP), transfer (backward/forward transfer, BWT/FWT), and forgetting (forgetfulness, F) of the models with an additional temporal dimension. We evaluate the models periodically on the frames held out from the same training video frames. The overall performance is aggregated from these evaluations and the transfer/forgetting is defined by the time intervals between the appearances of instances from the same tasks. We adapt several typical continual learning algorithms (e.g., iCaRL~\cite{rebuffi2017icarl}, EWC~\cite{kirkpatrick2017overcoming}, Incremental fine-tuning) to object detection and find the performance of these approaches mediocre in the new benchmark, which leaves substantial room for future studies.

\section{Related Work}
% In this section, we connect our work to the existing (online) continual learning benchmarks and settings.

\minisection{Continual learning benchmarks.} A large body of continual learning algorithms~\cite{aljundi2018memory,kirkpatrick2017overcoming,lopez2017gradient, schwarz2018progress,zenke2017continual, li2017learning, yoon2017lifelong, ruvolo2013ella, xu2018reinforced, javed2019meta, lee2020neural} have been developed and evaluated on image classification benchmarks such as Permuted MNIST~\cite{goodfellow2013empirical}, CIFAR-100~\cite{krizhevsky2009learning}, and  ImageNet~\cite{deng2009imagenet}. More recently,
Lomonaco and Maltoni~\cite{lomonaco2017core50} introduced CORe50, a collection of 50 domestic objects belonging to 10
categories, which supports image classification at
object level (50 classes) or at category level (10 classes) and object detection in a recent update.  In contrast to our work, the task splits in CORe50 are created manually and the benchmark is used for offline learning. 

In the object detection domain, several incremental object detection
algorithms~\cite{kuznetsova2015expanding, liu2020multi, perez2020incremental, shmelkov2017incremental} adopt existing object detection datasets such
as PASCAL VOC~\cite{Everingham15} and MS COCO~\cite{lin2014microsoft} for evaluation. They split the categories and train the object detectors on a pre-defined order of the categories sequentially. Chen~\etal.~\cite{Chen13} study the problem of continual learning by building NEIL with internet images. Kuznetsova~\etal~\cite{kuznetsova2015expanding} extend the evaluation from static images to video snippets like Activity Daily Living (ADL)~\cite{pirsiavash2012detecting} dataset and YouTube Objects (YTO) datasets~\cite{prest2012learning} for incremental domain adaptation. 

The task splits in these benchmarks are often manually determined and the classes
in the datasets are carefully balanced. Moreover,
existing benchmarks are mostly constructed using static images~\cite{deng2009imagenet,goodfellow2013empirical,krizhevsky2009learning} or samples from short
object-centric video snippets ($\sim$15 seconds)~\cite{lomonaco2017core50}. These existing benchmarks may not sufficiently unveil the challenges of building human-like continual learning agents given their artificial settings and static data sources. 

% Our dataset and benchmark are more realistic compared to the existing ones. We fully annotate an \emph{ego-centric view video stream} captured by
% a single person over 9 months, which enables the
% machine learning models to learn to perceive the world through the human's eyes and absorb a similar amount of
% visual information as humans do in their daily lives.
% Therefore, the tasks in our dataset follow a \emph{natural distribution} and an object category
% may be revisited as the person wanders around. Our data also follows a \emph{long-tail} distribution, reflecting the frequency of common and rare objects the person may encounter in their life. We see our dataset as a stepping stone towards building embodied agents that can continually learn new knowledge like humans. 

\begin{figure*}[t]
\centering
\includegraphics[width=0.9\linewidth]{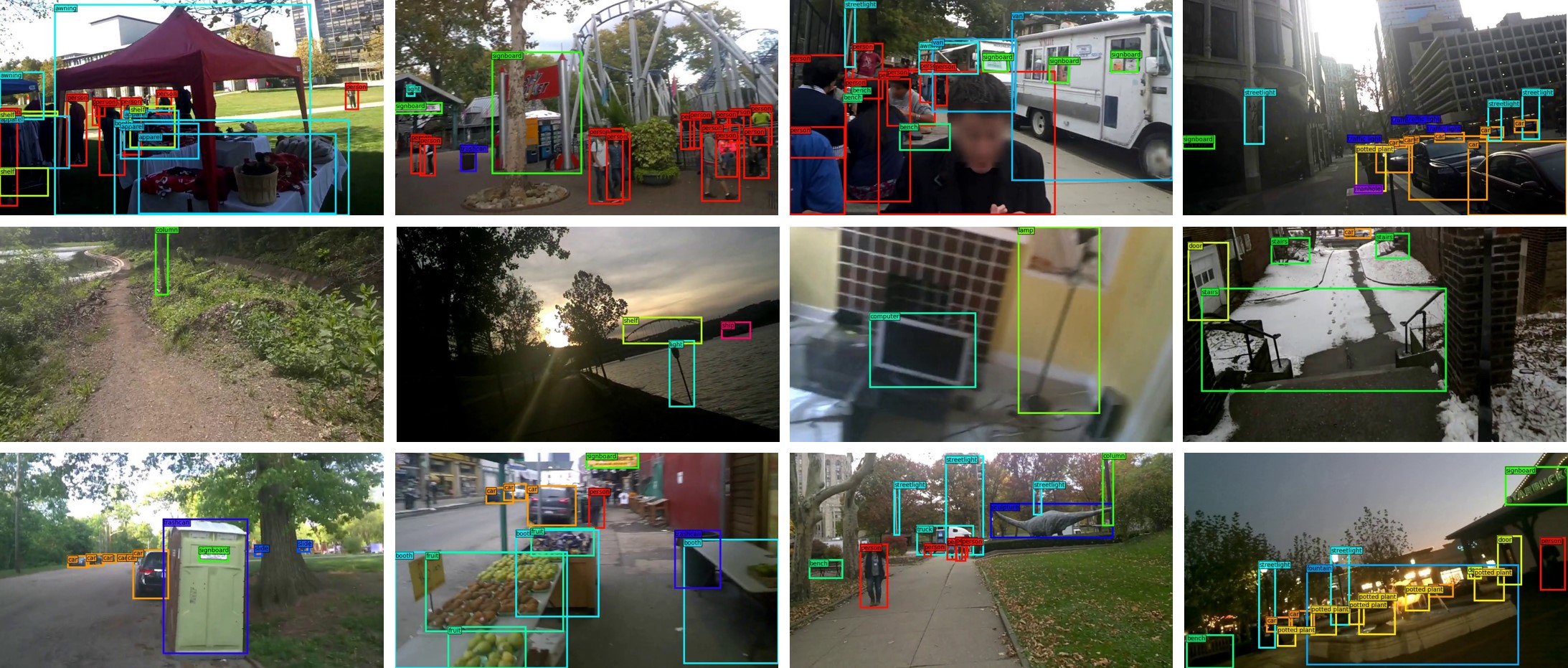}
\caption{Examples of annotated frames in \data. \data captures daily outdoor activities of a single graduate student. The dataset spans a wide variety of environments and life experiences.}

\label{fig:dataset_v}
\end{figure*}

\minisection{Online continual learning settings.} There is an emerging line of works on online continual learning~\cite{fini2020online,aljundi2019online,caccia2020online,aljundi2019task,ren2020wandering}. Aljundi~\etal~\cite{aljundi2019task} develop a system that keeps on learning over time in an online fashion, with data distributions gradually changing and without
the notion of separate tasks. Ren~\etal~\cite{ren2020wandering} recently extend the standard framework for few-shot learning to an online,
continual setting. Similar to Aljundi~\etal,  our online continual learning settings do not have an explicit task boundary. Since the tasks emerge as the person moves around, we introduce new metrics to evaluate the model transfer and forgetting through time.

\minisection{Ego-centric video recognition.} Another related research direction is ego-centric video recognition~\cite{ren2009egocentric, fathi2011learning, kapidis2019egocentric}. Most works focus on developing methods that are suitable for the unique perspective in an ego-centric video using an offline setting~\cite{yuan2019ego, kapidis2019egocentric}. To this sense, the most related work is ~\cite{erculiani2019continual}, where they tackle the problem of continual image classification. However, the dataset they use is~\cite{lomonaco2017core50} extremely short and clean, which cannot be sufficient to unveil the challenges in ego-centric video recognition. 

\section{\data Dataset}
\label{sec:dataset}
% In this section, we give an overview of the dataset construction and the data statistics. 
% We describe the dataset construction(Section~\ref{sec:annotation}) and provide statistical analysis of the dataset (Section~\ref{sec:data_stats}).

\begin{figure*}[t]
\centering
\subfloat[The number of categories increase over time but the emergence rate of novel categories decrease over time.\label{fig:da1}]{\includegraphics[width=.28\textwidth]{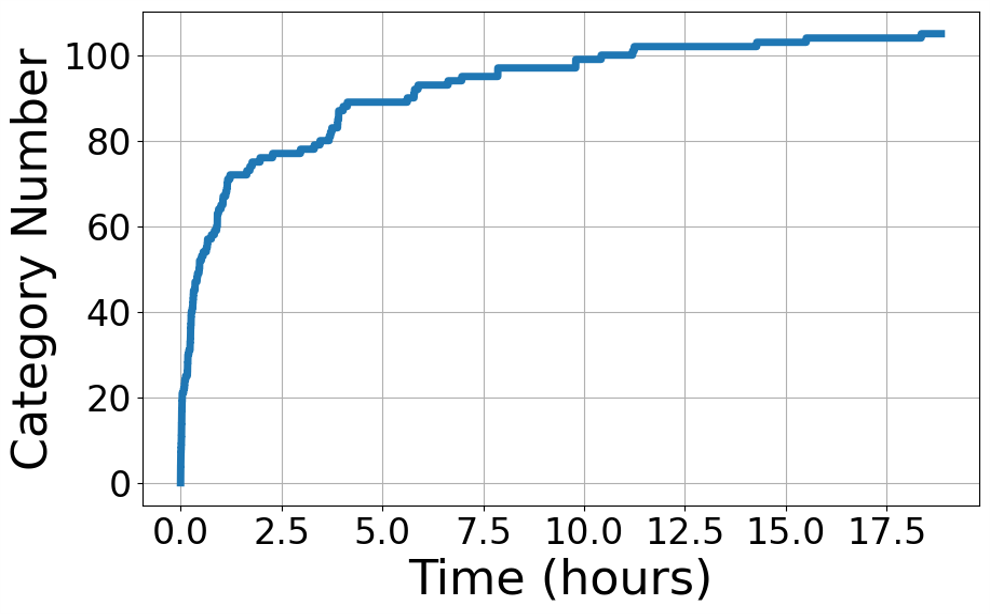}}
\hspace{2 mm}
\subfloat[The number of instances per category reveals a long tail distribution with existence of many rare classes.\label{fig:db1}]{\includegraphics[width=.28\textwidth]{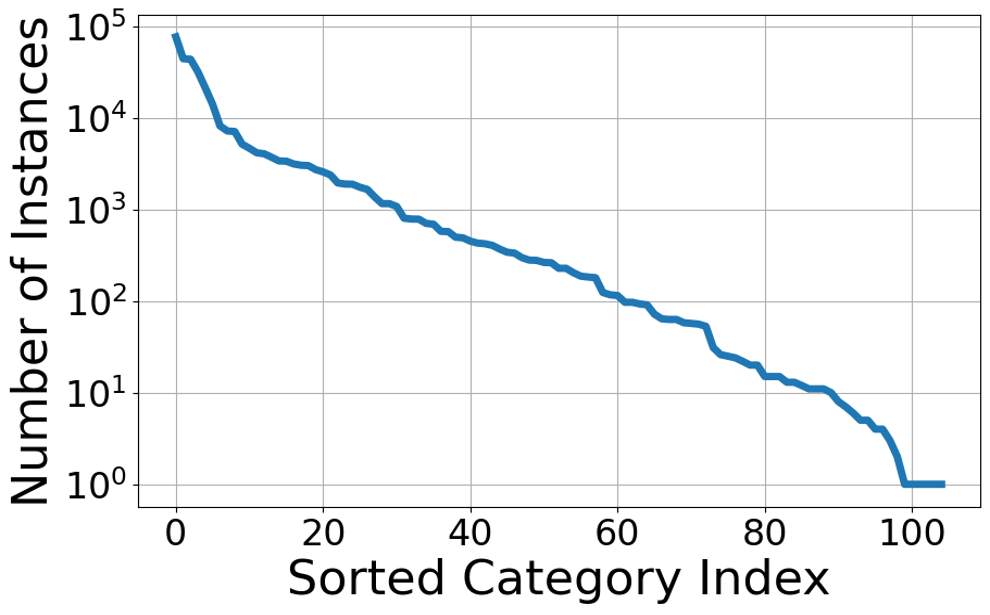}}
\hspace{2 mm}
\subfloat[Distribution of number of categories per image also reveals a long tail distribution. \label{fig:dc}]{\includegraphics[width=.28\textwidth]{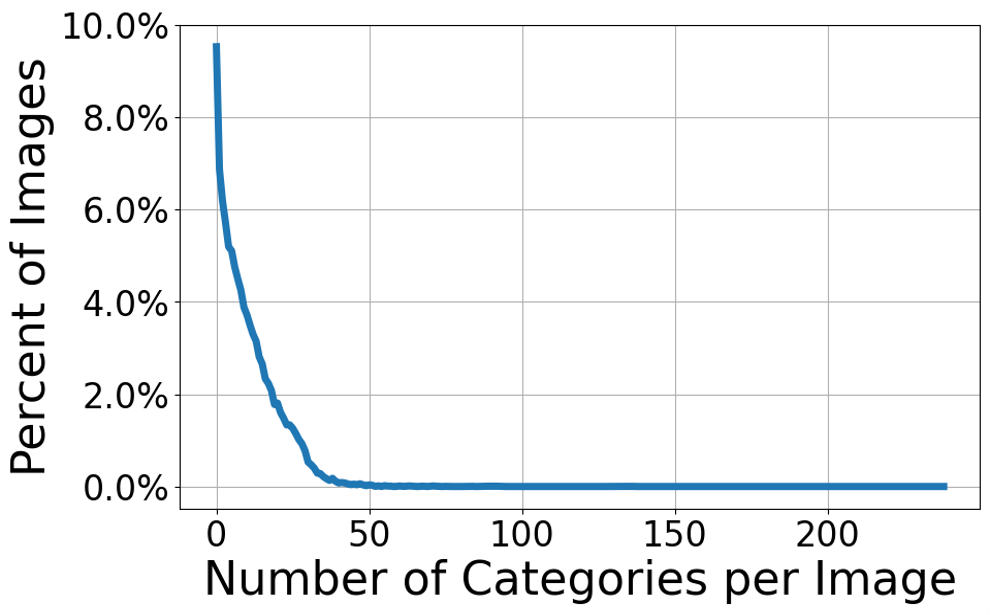}}

\subfloat[Distribution of time interval between the reappearances of data points from the same category. \label{fig:dd}]{\includegraphics[width=.28\textwidth]{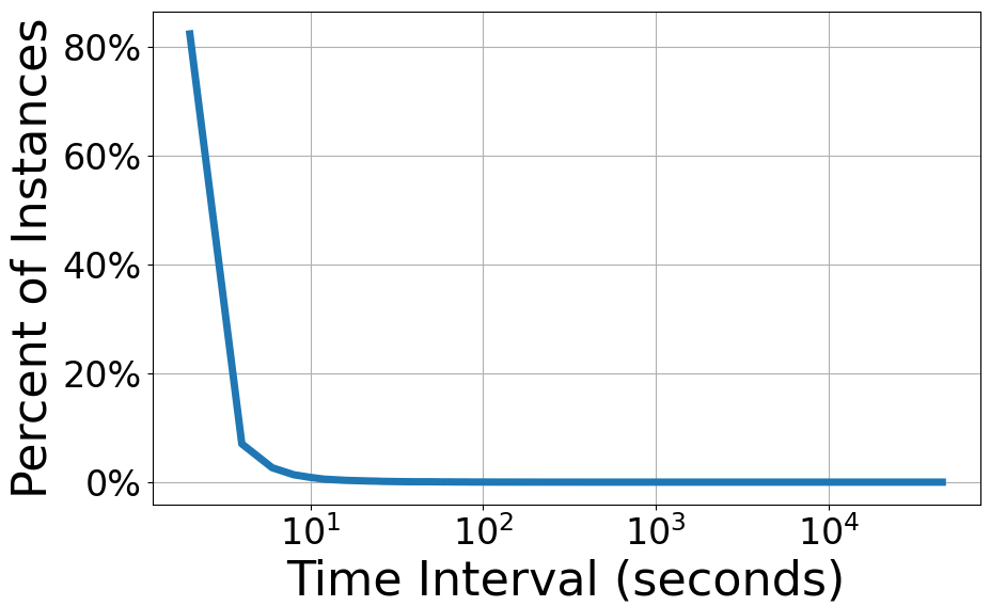}}
\hspace{2 mm}
\subfloat[Distribution of the bounding box size (pixel space). The distribution is skewed to small / medium size objects. \label{fig:de}]{\includegraphics[width=.28\textwidth]{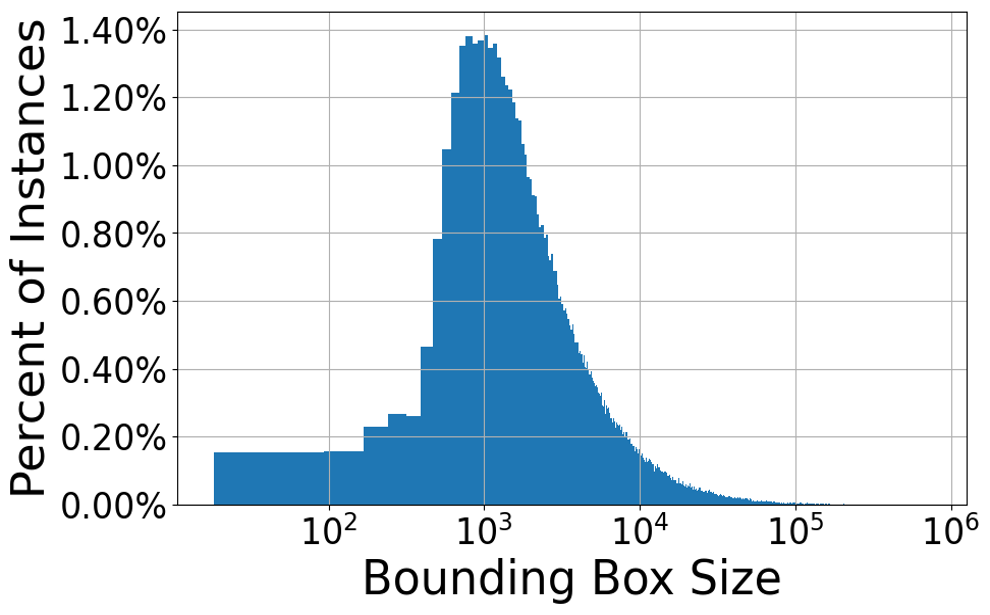}}
\hspace{4 mm}
\subfloat[Location distribution. Places in red are often visited and places in blue are occasionally visited.\label{fig:df}]{\includegraphics[width=.25\textwidth]{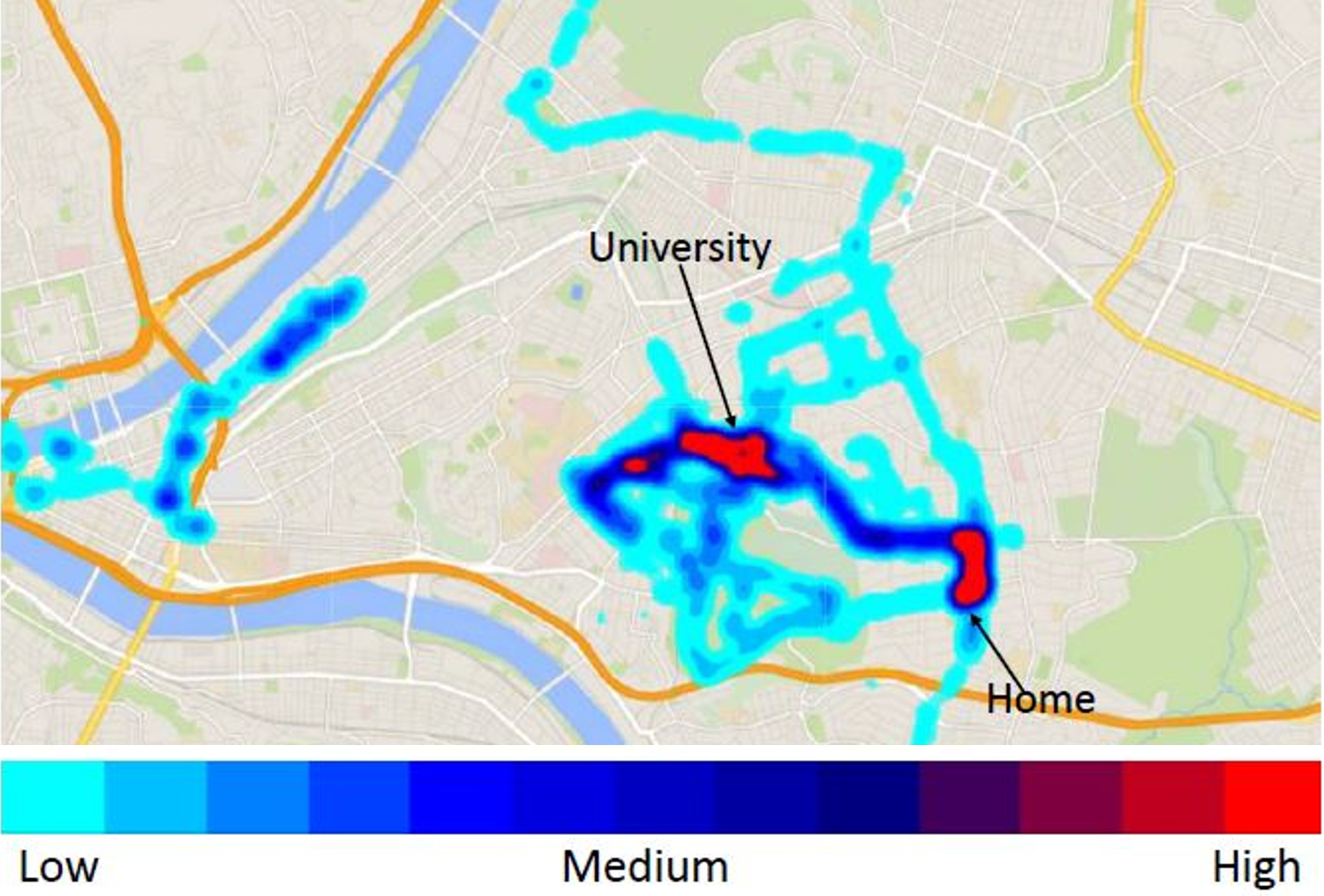}}
\vspace{-4mm}
\caption{Dataset statistics. Best viewed digitally.}
\label{fig:dataset_a}
\end{figure*}

% \subsection{Dataset Construction}
% \label{sec:annotation}
We introduce \emph{\datafull} (\data),
which labels the objects that appear as Krishna wanders and performs his daily routine. \data is built upon the \kcam~\cite{Singh16Krishn} dataset, a large
egocentric video stream spanning nine months of a graduate student's life. The original \kcam dataset contains 7.6 million frames of 460 video snippets with a total length of 70.2 hours. The raw videos have a resolution of 720p and a frame rate of 30 fps. In \data, we label around 1/4 of the original dataset and release the labeled data in the \hyperlink{https://oakdata.github.io/}{website}.

\minisection{Annotation setup.} We consider 105 object categories in outdoor scenes where 16 classes are from the PASCAL VOC dataset~\cite{Everingham15} and the remaining categories are frequent classes determined by running the LVIS~\cite{gupta2019lvis} pre-trained Mask R-CNN model~\cite{he2017mask} on the raw videos. The full list of the categories is provided
in the \hyperlink{https://oakdata.github.io/}{website}. We sample 80 videos from \kcam, which uniformly span in time. Each video snippet is about 7$\sim$15 minutes long. Objects of the 105 categories are exhaustively labeled at a frame rate of 0.5 fps. In total, \data contains roughly 326K bounding boxes. Two human annotators are involved in
labeling each frame to ensure the label quality. Objects in the frames are exhaustively annotated except for the tiny objects (less than 20$\times$ 20 pixels). In Figure~\ref{fig:dataset_v}, we show some examples of the annotated frames in \data. 

\minisection{Train and evaluation sets.} In contrast to the offline settings, the model is expected to train while
evaluating on the same data stream in an online continual learning setting to evaluate the catastrophic forgetting. Therefore, we hold out one
frame every 16 labeled frames to construct an
evaluation set and the remaining frames are used for training. The training and the evaluation sets cover a
similar time range of 9 months with different sample rates. The models will be trained and evaluated in an online manner. 

% \subsection{Dataset Statistics}
% \label{sec:data_stats}
\minisection{Dataset statistics.}
We show some data statistics in Figure~\ref{fig:dataset_a}.

\vspace{1mm}\noindent\emph{Natural task distribution.} In our online continual learning setting, a new task is defined as recognizing a new object category that was not encountered before. As shown in Figure~\ref{fig:da1}, the number of seen categories gradually increases and the appearance rate of novel classes decreases over time due to the repeated patterns in daily life.

\vspace{1mm}\noindent\emph{Long-tail distribution.} In Figure~\ref{fig:db1} and~\ref{fig:dc},  we can see \data has a long tail distribution both for the number of instances per category and the category counts per image. In Figure~\ref{fig:dd}, we show the distribution of time intervals between the reappearance of instances from the same category. 

\vspace{1mm}\noindent\emph{Skewed box sizes.} As shown in Figure~\ref{fig:de}, the box sizes in \data are skewed towards small and medium sizes, which makes it hard for the detector to make correct predictions. 

\vspace{1mm}\noindent\emph{Diverse geo-locations.} In Figure~\ref{fig:df}, we plot the locations of the video recordings. We can see that places like campus and home are frequently visited while other places are only occasionally visited. 

\section{Online Continual Learning Benchmark}
% We provide details about our benchmark setup in Section~\ref{sec:setting} and the evaluation metrics in Section~\ref{sec:metrics}.  

% \subsection{Online Continual Learning Settings}
% \label{sec:setting}
In this benchmark, we consider two online continual learning settings depending on whether the categories of interest are known beforehand. In both cases, the training data and labels emerge sequentially following the time stamps and the model is evaluated every N training steps. The main difference is the way that unseen categories are dealt with at the time of evaluation. If the model has a known vocabulary of the classes referred as the \texttt{known} setting, we can simply report the average precision (AP) on each category at each evaluation although some categories may not have been trained at the time of evaluation (often leading to a lower evaluation result).

The case where the model has an open vocabulary of classes, referred as the \texttt{unknown} setting,  is a bit more challenging but also more realistic. We introduce an IDK (I don't know) class for the unseen categories at the
current time stamp. For all the objects from the unseen categories in the evaluation set, the model needs to predict IDK for a correct prediction. The average precision of predicting IDK is also part of the evaluation protocol, which indicates the model's ability to identify new classes. For simplicity, we don't require the model to distinguish the exact category among the unseen categories as long as the model predicts IDK for the unseen objects.

\subsection{Evaluation Metrics}
\label{sec:metrics}
In the online continual learning setting, we focus on three aspects of the learned model: how well does the model perform overall? How well does the model transfer new knowledge? How resistant is the model to catastrophic forgetting? To this end, we introduce five evaluation metrics: continual average precision (CAP) and final average precision (FAP) for overall performance evaluation; forward transfer (FWT) for transfer performance evaluation; backward transfer (BWT) and forgetfulness (F) for forgetting performance evaluation. It's worth noting that transfer and forgetfulness are more comparable if two models have similar CAP. We adopt the commonly used AP50 (i.e., the average precision score with an IoU threshold of 50\%) in object detection for measurement. 

\minisection{CAP} shows the overall performance of the model in the time span of the entire video stream. Inspired by OSAKA~\cite{caccia2020online}, the accuracy of each timestep is evaluated using the current model instead of the final model. At each time $t$, the model is trained using a small batch ($t \times b$-th training frame to $(t+1) \times b$-th training frame, where $b$ denotes batch size) of data from $\mathcal{D}_\text{train}$. After time $t$, this small batch of images is no longer allowed to be used. We continue this training process until the entire video stream is covered. Every $N$ training steps, the model is evaluated on the test set $\mathcal{D}_\text{test}$. The reported \texttt{CAP}$_{t_i}$ ($i^{th}$ evaluation step) is defined as 
\begin{equation}
    \texttt{CAP}_{t_i} = \frac{1}{C}\sum_{c=0}^{C} \texttt{CAP}_{t_i}^c,
\end{equation}
where $\texttt{CAP}_{t_i}^c$ is the average precision (AP) of the class $c$ on the 
test set. \texttt{CAP} is then defined as the average values across different time stamps. That is, 
\begin{equation}
    \texttt{CAP} = \frac{1}{T}\sum_{i=0}^T \texttt{CAP}_{t_i} = \frac{1}{TC}\sum_{i=0}^T\sum_{c=0}^C \texttt{CAP}_{t_i}^c,
\end{equation}
where $T$ is the total evaluation times.

\minisection{FAP} is the final average precision of the last model when the model finishes training. This is a more fair evaluation metric when comparing with offline learning models as both models have observed the entire video stream. 

% \minisection{GAP} evaluates the performance gain after the model is trained on the new data point to estimate the learning speed. For a class $c$, when the model is trained on new data of $c$ at time $t$, the performance gain is defined as 
% \begin{equation}
%     \texttt{GAP}_t^c =\texttt{CAP}_t^c-\texttt{CAP}_{t-1}^c.
% \end{equation}

% Since the data points of the class $c$ may appear multiple times in the video stream, we report the average value of $\texttt{GAP}_t^c$ as 
% \begin{equation}
%     \texttt{GAP}^c = \frac{1}{T_c}\sum_{t=0}^T \mathbb{I}[c\in\mathcal{D}_\text{train}^t]\texttt{GAP}_t^c,  
% \end{equation}
% where $T_c$ is the number of appearances of the class $c$. The overall performance gain is defined as 
% \begin{equation}
%     \texttt{GAP} = \frac{1}{C}\texttt{GAP}^c.
% \end{equation}

\minisection{FWT} evaluates the forward transfer ability of new knowledge inspired by GEM~\cite{lopez2017gradient}. FWT shows the influence that learning a scenario (a video clip, denoted as $\mathcal{S}_\text{t}$) has on the performance for future scenarios ($\mathcal{S}_\text{k}, k > t$). The scenarios are a short clip with a fixed interval of 0.94 hour in our benchmark. Positive forward transfer is possible when the model is able to perform “zero-shot” learning. Specifically, we divide both $\mathcal{D}_\text{train}$ and $\mathcal{D}_\text{test}$ evenly into $T$ scenarios in temporal order, where each division of $\mathcal{D}_\text{train}^i$ and $\mathcal{D}_\text{test}^i$ is used as train set and test set of scenario $\mathcal{S}_\text{i}$. 

After the model finishes learning from scenario $\mathcal{S}_\text{i}$, we evaluate its test performance on all $T$ scenarios. By doing so, we
construct the matrix $R\in \mathcal{R}^{T \times T}$, where $R_{i,j}$ is the test mean average precision (mAP) of the model on scenario $\mathcal{S}_\text{j}$ after observing the last frame from $\mathcal{S}_\text{i}$. Letting $\Bar{b}$ be the vector of test mAP of an pre-trained object detector for each scenario, we define FWT as:

\begin{equation}
    \texttt{FWT} = \frac{1}{T-1}\sum_{i=2}^T R_{i-1,i}-\Bar{b}_i.
\end{equation}

\minisection{BWT} shows the influence that learning a scenario (a video clip, denoted as $\mathcal{S}_\text{t}$) has on the performance on previous scenarios ($\mathcal{S}_\text{k}, k < t$). Negative backward transfer is also known as forgetting. Specifically, we define BWT as:

\begin{equation}
    \texttt{BWT} = \frac{1}{T-1}\sum_{i=1}^{T-1} R_{T,i}-R_{i,i}.
\end{equation}

\minisection{Forgetfulness (F)} estimates the model forgetting due to the sequential training. For a class $c$, we sort the $\texttt{CAP}_{t_i}^c$ according to the time interval $k$ between evaluation time $t_i$ and the last time $t_i-k$ the model is trained on $c$. After the $\texttt{CAP}_{t_i}^c$ is sorted, all $\texttt{CAP}_{t_i}^c$ ($i=0,\dots,T$) are divided into $K$ bins $B_{kmin},...,B_{kmax}$ according to the time interval $k$. The average $\texttt{CAP}$ ($\texttt{aCAP}_k$) of each bin $B_k$ is defined as the model's performance for detecting class $c$ after the model have not been trained on $c$ for $k$ time stamps. We define forgetfulness (F) of the class $c$ as the weighted sum of the performance decrease at each time:

\begin{equation}
    \texttt{F}^c = \sum_{k=kmin}^{kmax} \frac{k-kmin}{\sum_{k=kmin}^{kmax} k-kmin} \times (\texttt{aCAP}_{kmin} - \texttt{aCAP}_k).
\end{equation}

Thus, the overall forgetting is defined as 
\begin{equation}
    \texttt{F} = \frac{1}{C}\sum_{c=0}^C \texttt{F}^c.  
\end{equation}

% \minisection{Forgetfulness (F)} estimates the model forgetting due to the sequential training. For a class $c$, we consider a window between $t_i^c$ and $t_{i+1}^c$ which are time stamps of the $i$ and $i+1$ appearance of the data points of $c$. At each time window, the forgetting is defined as the weighted sum of the performance decrease at each time. That is, 
% \begin{equation}
%     \texttt{F}_i^c = \sum_{t=t_i^c}^{t_{i+1}^c} \alpha_t^i \left( \texttt{CAP}_{t_i^c}^c - \texttt{CAP}_t^c\right), 
% \end{equation}
% where $\alpha_t^i = (t-t_i^c)/(t_{i+1}^c - t_i^c)$. The overall forgetting of the class $c$ is the average across multiple windows as the data points of the class $c$ may appear multiple times. Thus, the overall forgetting is defined as 
% \begin{equation}
%     \texttt{F} = \frac{1}{C}\sum_{c=0}^C \texttt{F}^c = \frac{1}{C}\sum_{c=0}^C\frac{1}{N_c}\sum_{i=0}^{N_c}\texttt{F}_i^c,  
% \end{equation}
% where $N_c$ is the appearances of the data points of $c$.

\section{Experiments}
We describe three widely adopted continual learning algorithms in Section~\ref{sec:baseline} and present their performance on our benchmark in Section~\ref{sec:res}. Although the existing continual learning algorithms generally improve over the non-adaptive model, the \texttt{CAP} value for each of the algorithms is less than 20. This indicates our benchmark is challenging, which has a large room for future algorithm designs.

\subsection{Continual Learning Algorithms}
\label{sec:baseline}
We select the three representative continual learning methods according to Parisi ~\textit{et al.}~\cite{parisi2019continual}. Incremental finetuning is intuitive and widely adopted as a baseline for continual learning. iCaRL~\cite{rebuffi2017icarl} is a widely used memory-based method. EWC~\cite{kirkpatrick2017overcoming} is a representative regularization-based method. We first show how each method is deployed to the \texttt{known} setting, and we then show that these methods can easily adapt to the \texttt{unknown} setting.

\begin{figure}[t!]
\centering
\includegraphics[width=0.9\linewidth]{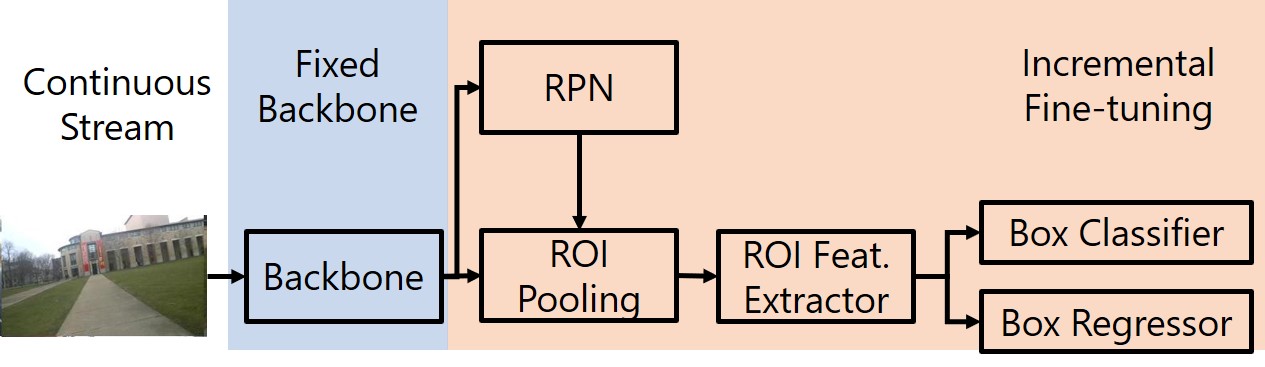}
\caption{Incremental fine-tuning. The entire object detector, including both the feature extractor $\mathcal{F}$ and the box predictor ($\mathcal{R}$ \& $\mathcal{C}$) are pretrained using PASCAL VOC. In online continual learning, the backbone is fixed, while RPN and box predictor are fine-tuned. \vspace{-4mm}}
\label{fig:baseline_finetune}
\end{figure}

\minisection{Incremental fine-tuning.} The first baseline is incremental fine-tuning on the widely used two-stage object detector, Faster R-CNN~\cite{ren2015faster}, which is pretrained on the PASCAL VOC dataset. As shown in Figure~\ref{fig:baseline_finetune}, the feature learning components $\mathcal{F}$ include the backbone (e.g., ResNet~\cite{he2016deep},  VGG16~\cite{Simonyan15}), the region proposal network (RPN), as well as a two-layer fully-connected (FC) sub-network as a proposal-level feature extractor. There is also a box predictor composed of a box classifier $\mathcal{C}$ to classify the object categories and a box regressor $\mathcal{R}$ to predict the bounding box coordinates. The backbone features, as well as the RPN features, are class-agnostic. However, we find keeping the RPN updated during fine-tuning significantly improves the performance. Thus, we only keep the backbone fixed during the incremental fine-tuning process. Whenever new data comes in, we fine-tune the RPN, the box classifier, and the regressor on the new data.

% Based on this intuition, we propose to fix the feature extractor and continually fine-tune the box classifier and box regressor. As one can imagine, incremental fine-tuning will lead to fast incremental learning while suffers from catastrophic forgetting.

\begin{figure}
\centering
\includegraphics[width=\linewidth]{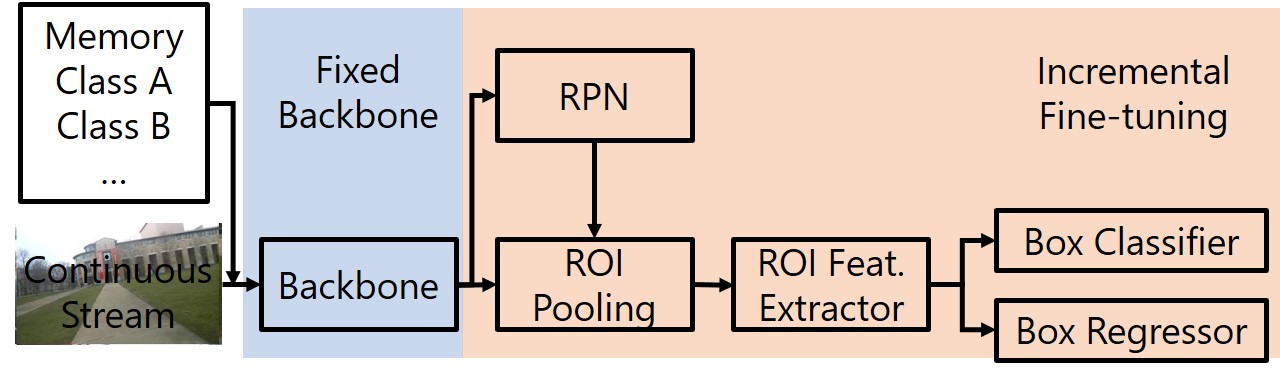}
\caption{Illustraion of iCaRL. The network parameters are updated by minimizing the loss on both streaming data and memorized prioritized exemplars.\vspace{-4mm}}
\label{fig:baseline_iCaRL}
\end{figure}

\minisection{iCaRL.} 
The second baseline adapts from the iCaRL~\cite{rebuffi2017icarl} algorithm proposed by Rebuffi~\etal. The original iCaRL is designed for image classification and we implement this approach with the Faster R-CNN based detector. As illustrated in Figure~\ref{fig:baseline_iCaRL}, iCaRL additionally includes a memory bank $\mathcal{M}$ to store the representative examples of each category, which are selected and updated randomly during each step. The memory bank has a fixed size (we set it to 5 images per category) and the old examples in the memory bank are replaced with newer data points. For each training step, we would randomly select a sample (image with one object label) from each class of the memory bank to train jointly. Same as incremental fine-tuning, we keep the backbone fixed and fine-tune the RPN, the box classifier, and the regressor when new data comes in.

\begin{figure}[t]
\centering
\includegraphics[width=0.9\linewidth]{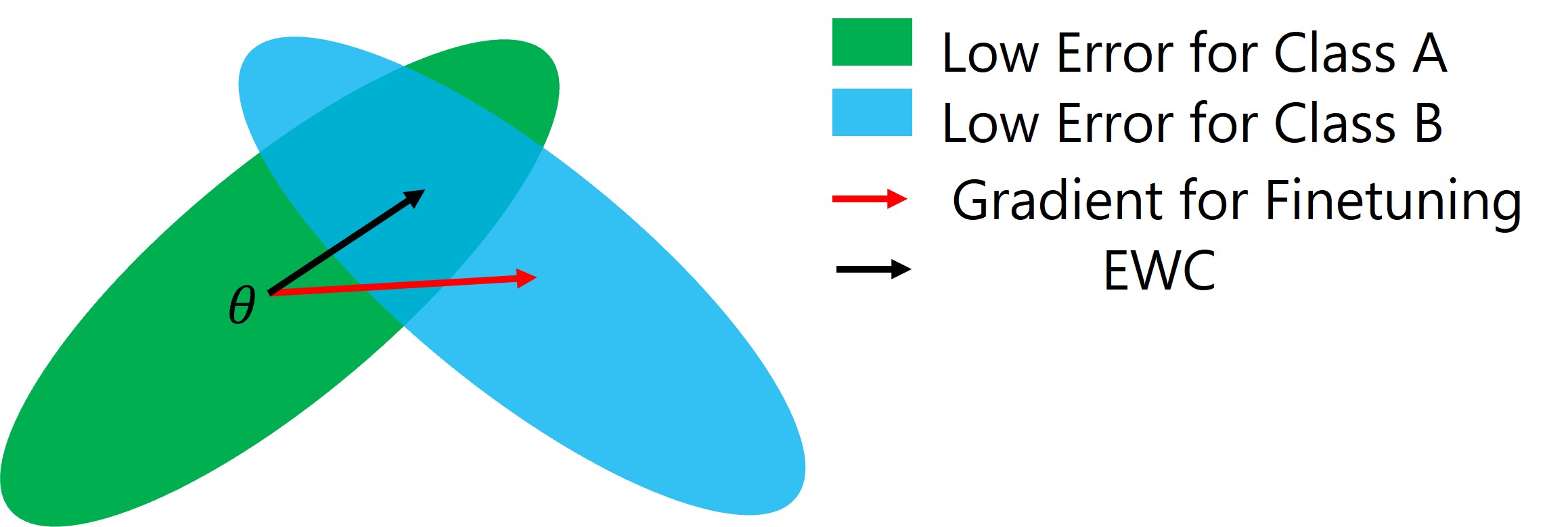}
\caption{Illustration of EWC (figure adapted from~\cite{lopez2017gradient}). When learning to predict a new class B, the gradient updates may hurt the performance of the old class A. EWC minimizes the classification loss without incurring a significant loss on the old classes.\vspace{-4mm}}
\label{fig:baseline_ewc}
\end{figure}

\minisection{EWC.} The third baseline adapts from the EWC algorithm~\cite{kirkpatrick2017overcoming}, proposed by Kirkpatrick~\etal. Similar to iCaRL, we apply the EWC algorithm to the box classifiers in Faster R-CNN. EWC does not require access to ground truth labels in the memory bank. The main idea of EWC is to impose constraints over the gradient updates so that the gradient updates on the new examples do not increase the classification loss on the old examples as illustrated by Figure~\ref{fig:baseline_ewc}. Interested readers can refer to the EWC paper~\cite{kirkpatrick2017overcoming} for detailed mathematical formulations.

\minisection{Other baselines.} We provide the performance of the Faster R-CNN pretrained on the PASCAL VOC data, denoted as \emph{non-adaptation} in the result tables. We also provide the model performance using offline training. At each evaluation time $t_i$ ($i^{th}$ evaluation step), the entire video stream before $t_i$ are used as one training set. We then conduct batch training offline and report the $\texttt{CAP}_{t_i}$ on $\mathcal{D}_\text{test}$. This baseline is denoted as \emph{offline training} in the result tables. 

% \minisection{Non-adaptation.} In the non-adaption baseline, the feature extractor $\mathcal{F}$, box classifier $\mathcal{C}$ and box regressor $\mathcal{R}$ of the pretrained object detector (i.e., Faster RCNN) are all fixed. In other words, the agent does not adapt the streaming data at all. This can be treated as the lower bound of online continual object detection.  

% \minisection{Offline training.} In the offline training baseline, the feature extractor $\mathcal{F}$ is fixed. While the box classifier $\mathcal{C}$ and box regressor $\mathcal{R}$ of the object detector (i.e., Faster RCNN) are fine-tuned offline using all labels from the training set in an offline manner. This is the standard train/test paradigm and should provide insight into the hardness of the dataset.

\minisection{Training details.}
For a fair comparison, we use ResNet-50~\cite{he2016deep} as the backbone of all continual learning algorithms, and the base object detector, Faster R-CNN, is pretrained on PASCAL VOC~\cite{Everingham15}.  \data share the same 20 categories with the PASCAL VOC dataset and thus the pretrained model can be used as an initial point. In the \texttt{known} setting, the number of categories in the detector is fixed and set to 105. 
In the \texttt{unknown} setting, the number of categories increases through time. We add a column to the box classifier and a box regressor when new data contains previous unseen categories. At the prediction time, if the confidence of all classes is smaller than a threshold while the confidence of background also maintains a low status, the agent should predict IDK. 

% we change the Softmax head~\cite{goodfellow2016deep} of incremental fine-tuning and EWC into multiple Sigmoid head~\cite{goodfellow2016deep}. Thus, the class predictions are decoupled from each other. 

% Experimentally, we set the background threshold to and calss 

\subsection{Overall Model Performance}
\label{sec:res}

\begin{table*}[t]
\centering
\caption{Overall performance of existing algorithms on \data measured by continual average precision (CAP) and final average precision (FAP) with known class vocabulary (\texttt{known}). FAP is consistently better than CAP, which shows all algorithms benefit from learning from more data. Due to the distribution shift, the non-adaptation model has the lowest performance. The regularization-based EWC approach performs similar to vanilla incremental fine-tuning. While the memory-based iCaRL approach is significantly better than vanilla incremental fine-tuning. The offline training results show that \data is a bit challenging to the current object detector even under the offline training setting.\vspace{-2mm}}
\label{tab:cap_known}
\adjustbox{width=\linewidth}{
\begin{tabular}{lcccccccccccc}
\toprule
Method & FAP & CAP & Top-20 & Booth & Umbrella &Awning & Bag & Chair & Dining Table & Fireplug & Car\\
\midrule
Non-adaptation & 2.86 & 2.86 & 11.72 & 0.00 & 0.00 & 0.00 & 0.00 & 23.88 & 26.38 & 0.0 & 55.90\\
\midrule
Incremental & 12.47 & 10.47  & 33.84 & 0.20 & 2.46 & \textbf{4.58} & \textbf{18.24} & \textbf{31.70} & 36.51 & 39.53 & 69.68\\
EWC & 12.55 & 10.52 & 33.80 & 0.21 & 3.02 & 4.53 & 17.79  & 31.38 & 36.67 & 38.73 &  \textbf{69.71}\\
iCaRL & \textbf{21.89} & \textbf{16.39} & \textbf{40.92} & \textbf{6.88} & \textbf{6.91} & 3.59 & 10.67 & 28.72 & \textbf{40.59} & \textbf{45.18} & 67.18\\
\midrule
Offline training & 49.48 & 37.11  & 67.77 & 20.05 & 35.86 & 28.65 & 42.69 & 66.51 & 71.09 & 64.15 & 78.80\\
\bottomrule
\end{tabular}}
\end{table*}

\begin{figure*}[t]
\centering
\includegraphics[width=0.9\textwidth]{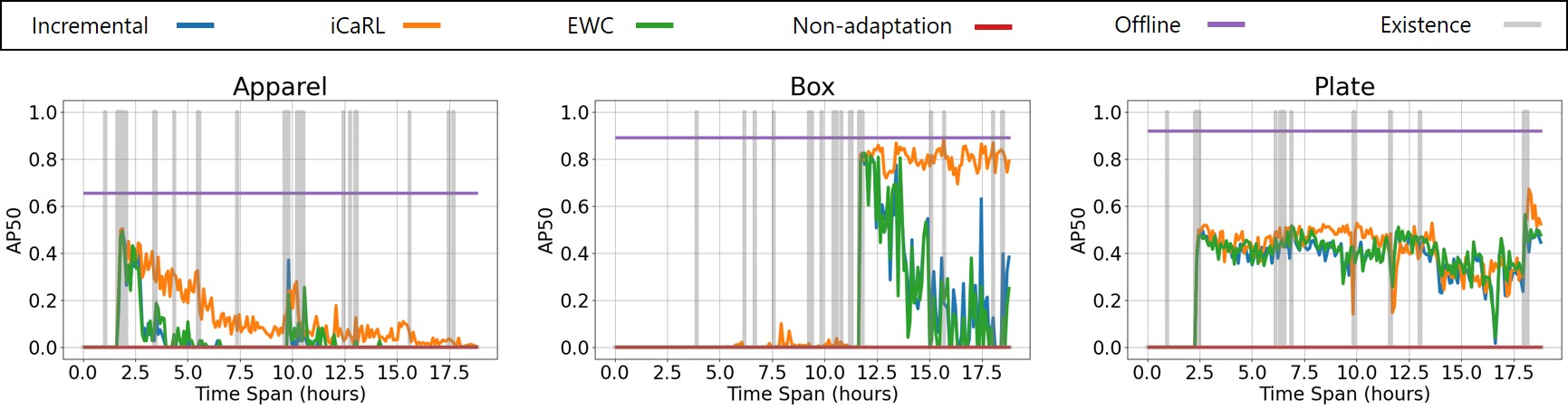}
\caption{$CAP_{t_i}$ changes on OAK test set, in chronological order. $CAP_{t_i}$ increases as the model is trained on frames where each category exists, but as more time passes between examples, the forgetting effect is observed and the performance falls.}
\label{fig:result_CAP}
\end{figure*}

\minisection{Known class vocabulary.} In Table~\ref{tab:cap_known}, we provide the overall model performance measured by continual average precision (\texttt{CAP}) and final average precision (\texttt{FAP}) under the \texttt{\texttt{known}} setting, where the vocabulary of the classes is known beforehand. The pretrained model on PASCAL VOC (Non-adaptation) only achieves 2.86 points on the new \data dataset. Even for the classes that overlap with the original PASCAL VOC data (e.g., chair, dining table, etc.), the performance is extremely low. This indicates that \data has a large domain gap with the existing detection dataset, which uses static Internet images. 

For the three continual learning algorithms, the memory-based approach iCaRL outperforms the vanilla incremental fine-tuning. This demonstrates that the simple rehearsal strategy can still play an important role in the new online continual learning setting. However, the regularization-based approach EWC does not help with the overall
performance in our setting, which might be due to the complication of the mixed task orders. It is also worth noticing that both EWC and iCaRL have a significant performance gap compared with the offline training. This indicates that the new online continual learning setting may require more innovation in the algorithm designs to deal with the new challenges posed by online learning. Additionally, since both of them are not designed for object detection, the model performance may be improved if one can design models specialized for object detection. 

We notice that the model performance of the offline training is only about 49.48 points. It is a relatively low model performance compared with the model performance on other detection datasets such as PASCAL VOC or COCO where the AP50 scores are often above 50 points. This gap is also discovered by other egocentric video challenges~\cite{damen2018scaling, ren2009egocentric}. We conjecture the new challenges posed by \data comes from the intrinsic features of ego-centric videos, such as motion blur, lots of occlusion and partially observed objects due to limited field of view and the long tail distribution.

\begin{table*}[t!]
\centering
\caption{Overall performance of the existing algorithms on \data measured by continual average precision (CAP) and final average precision (FAP) with an unknown class vocabulary (\texttt{unknown}). In this evaluation, the model needs to predict IDK for the objects in the unseen categories. All the continual learning algorithms considered have a decreased performance compared with the \texttt{known} setting. Again Incremental finetuning and EWC achieve similar performance. iCaRL outperforms incremental fine-tuning and EWC but there is substantial room for improvement.\vspace{-2mm}}
\label{tab:cap_uk}
\adjustbox{width=\linewidth}{
\begin{tabular}{lcccccccccccc}
\toprule
Method & FAP & Overall & Top-20 & Booth & Umbrella &Awning & Bag & Chair & Dining Table & Fireplug & Car & IDK\\
\midrule
Non-adaptation & 2.86 & 2.86 & 11.72 & 0.00 & 0.00 & 0.00 & 0.00 & 23.88 & 26.38 & 0.0 & 55.90 & -\\
\midrule
Incremental & 11.19 & 9.78 & 32.10 & 0.06 & 1.45 & \textbf{3.87} & \textbf{16.68} & 29.26 & 32.81 & 36.32 &  68.76 & 0.32\\
EWC & 10.34 & 9.57 & 31.62 & 0.21 & 1.63 & 3.72 & 16.46 & \textbf{29.76} & \textbf{33.39} & 33.41 & \textbf{68.76} & 0.29\\
iCaRL & \textbf{17.56} & \textbf{14.70} & \textbf{41.42} & \textbf{3.58} & \textbf{4.83} & 2.72 & 10.18 & 23.81 & 32.86 & \textbf{39.84} & 64.02 & \textbf{0.36}\\
\midrule
Offline training & 46.40 & 36.88  & 71.35 & 18.90 & 31.90 & 29.36 & 41.36 & 64.59 & 70.57 & 63.77 & 78.36 & 1.21\\
\bottomrule
\end{tabular}}
\end{table*}

We visualize the CAP for each class during online continual learning. As shown in Figure~\ref{fig:result_CAP}, the shading indicates when the agent observes new data points from a specific class, while the points on each curve are evaluation steps on OAK. Notice how in each class the performance increases initially, but as more time passes between examples, the forgetting effect is observed and the performance falls. When the next round of data is seen towards the end, the performance improves again.

\minisection{Unknown class vocabulary.} In Table~\ref{tab:cap_uk}, we provide the overall performance of the algorithms with an unknown class vocabulary measured by CAP and FAP. In this evaluation, the model needs to predict IDK for the objects in the unseen categories, which is more challenging than the \texttt{\texttt{known}} setting. As we can see from the table, all the continual learning algorithms considered have a decreased performance compared with the \texttt{known} setting. Again iCaRL outperforms incremental fine-tuning and EWC but there is substantial room for improvement.

% \begin{table*}[t]
% \centering
% \caption{Learning speed measured by gain average precision (GAP) in the \texttt{known} setting. GAP measures 
% the difference of CAP values before and after the model is trained on the new data. Higher GAP indicates a faster learner. Incremental fine-tuning has a higher performance gain after training compared with GEM and iCaRL. iCaRL is more conservative about learning from new data points, with negative values for a few categories. 
% \vspace{-2mm}}
% \adjustbox{width=.95\linewidth}{
% \begin{tabular}{lccccccccccc}
% \toprule
% Method & Overall & Top-20 & Street light & Bag & Awning & Chair & Booth & Fireplug & Dining Table & Car\\
% \midrule
% Incremental & \textbf{1.07} & \textbf{1.73} & 0.02 & 0.08 & -0.29 & 0.10 & \textbf{0.52} & 0.36 & 0.51 & \textbf{0.03}\\
% iCaRL & -2.04 & -0.18 & -0.34 & -0.88 & -0.64 & -0.96 & -2.35 & \textbf{0.51} & 0.27 & -0.40\\
% GEM & 0.16 & 0.63 & \textbf{0.02} & \textbf{0.22} & \textbf{-0.18} & \textbf{0.11} & -0.70 & 0.25 & \textbf{0.65} & -0.04\\
% \bottomrule
% \end{tabular}}
% \label{tab:gap}
% \end{table*}

\begin{table*}[t]
\centering
\caption{Forgetfulness of the continual learning algorithms. The forgetfulness (F) metric indicates the catastrophic forgetting and smaller F scores mean less forgetting on the learned knowledge. iCaRL is better at avoiding catastrophic forgetting than the incremental fine-tuning and EWC. \vspace{-2mm}}
\adjustbox{width=.9\linewidth}{
\begin{tabular}{lccccccccccc}
\toprule
Method & Overall & Worst-20 & Booth & Umbrella & Awning & Bag & Chair & Dining Table & Fireplug & Car\\
\midrule
Incremental & 5.77 & 22.28 & 0.60 & \textbf{1.10} & 1.28 & -1.95 & 2.91 & \textbf{-0.29} & 11.89 & 5.75\\
EWC & 5.96 & 23.48 & 0.43 & 1.85 & 1.13 & \textbf{-2.00} & 2.76 & 0.02 & 10.30 & 5.76\\
iCaRL & \textbf{1.78} & \textbf{14.72} & \textbf{-5.65} & 2.03 & \textbf{0.67} & -1.41 & \textbf{1.50} & 0.68 & \textbf{6.79} & \textbf{3.32}\\
\bottomrule
\end{tabular}}
\label{tab:fap}
\end{table*}

\begin{figure*}[t!]
\centering
\includegraphics[width=0.9\textwidth]{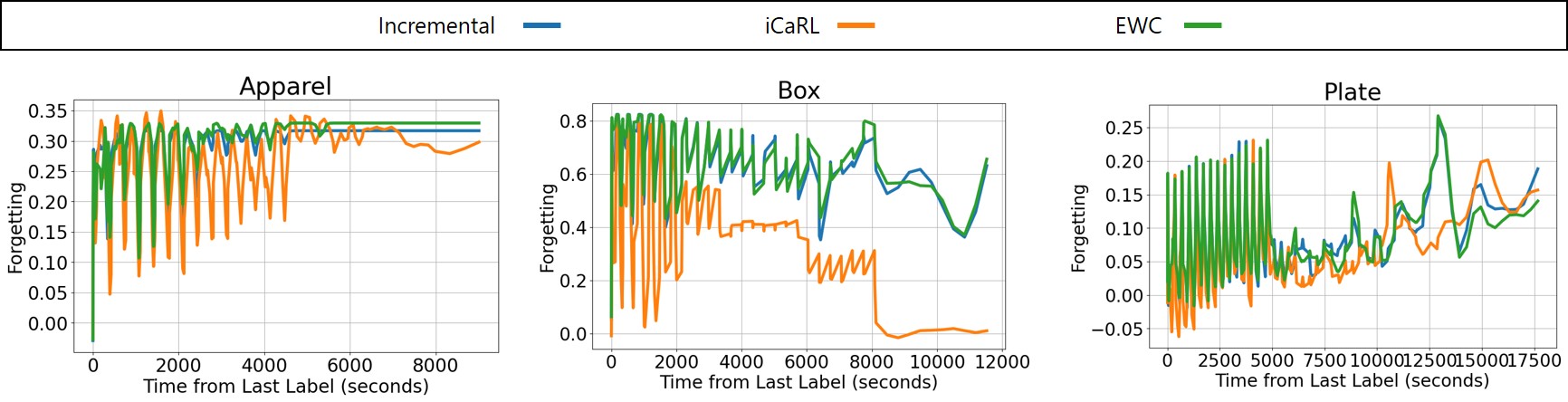}
\caption{Forgetfulness of sample categories. We provide the evaluation curve on the test set under different time stamps.}
\label{fig:result_F}
\end{figure*}

\subsection{Transfer and Forgetting}

\begin{table}[t]
\centering
\adjustbox{width=.5\linewidth}{
\begin{tabular}{lcc}
\toprule
Method & FWT & BWT\\
\midrule
Incremental & 15.01 & -4.51\\
EWC & 14.84 & \textbf{-3.62}\\
iCaRL & \textbf{15.73} & -5.95\\
\bottomrule
\end{tabular}}
\caption{Knowledge transferability under the \texttt{known} setting. Incremental fine-tuning has better forward transferability and worse backward transferability. This shows that gradient regularization can alleviate catastrophic forgetting at the cost of information gain.
}
\label{tab:fwt_bwt}
\end{table}

\minisection{Transfer.} In the first column of Table~\ref{tab:fwt_bwt}, we compare the forward transferability of three continual learning algorithms under the \texttt{known} setting. Higher FWT indicates a faster learner. Incremental fine-tuning has higher FWT compared with EWC. EWC is more conservative about learning from new data points with regularization of the weight change. We don't compare transferability for \texttt{unknown} setting since the tasks keep changing.

\minisection{Forgetting.} In Table~\ref{tab:fap}, we present the forgetfulness of the continual learning algorithms under the \texttt{known} setting. The forgetfulness (F) metric indicates the amount of catastrophic forgetting. Smaller F scores indicate less forgetting on the learned knowledge. iCaRL is significantly better at avoiding catastrophic forgetting than the incremental fine-tuning method and EWC. For more comparison with existing metrics, we also compare the backward transferability of three continual learning algorithms under the \texttt{known} setting (second column of Table~\ref{tab:fwt_bwt}).  Higher BWT indicates less forgetting. EWC has higher BWT than Incremental fine-tuning since EWC finds solutions for new tasks without incurring significant losses on old tasks. We choose forgetfulness (F) as our main metric to estimate the model's forgetting due to the sequential training.

We also visualize the forgetfulness changes of each class during online continual learning. As shown in Figure~\ref{fig:result_F}, the forgetfulness increases as the time from seeing the last labeled data increases. Notice that in most cases, the incremental fine-tuning baseline performs best.

\section{Conclusion}
Online continual learning from continuous data streams in dynamic environments is of an increasing interest in the machine learning and computer vision community. However, realistic datasets and benchmarks, especially for object detection, are still missing. 
In this work, we presented a new online continual object detection benchmark dataset called OAK. OAK uses the videos from the KrishnaCAM dataset, which features an ego-centric video stream collected over a nine-month time span. OAK provides exhaustive annotations of 80 video snippets ($\sim$17.5 hours) with 326K bounding boxes of 105 object categories in outdoor scenes. Our continual learning benchmark follows the life pattern of a single person, which is more realistic compared to the existing continual learning benchmarks. We introduced new evaluation metrics as well as baseline evaluations for two evaluation setups. We hope this work inspires research in embodied online continual learning of object detectors.

\minisection{Acknowledgements.} The authors would like to thank Yi Ru Wang, Samantha Powers, Kenneth Marino, Shubham Tulsiani for fruitful discussion and detailed feedback on the manuscript. Carnegie Mellon Effort has been supported by DARPA MCS, ONR Young Investigator, ONR muri.

{\small
\bibliographystyle{ieee_fullname}
\bibliography{references}
}

\end{document}

%%%%%%%%% TITLE
\title{Supplementary Materials\\
Wanderlust: Online Continual Object Detection in the Real World}

\author{First Author\\
Institution1\\
Institution1 address\\
{\tt\small firstauthor@i1.org}
\and
Second Author\\
Institution2\\
First line of institution2 address\\
{\tt\small secondauthor@i2.org}
}

% \maketitle
% % Remove page # from the first page of camera-ready.

\maketitle

\section{Dataset Details}

In this section, we provide more details about labeled categories. Supplementary Table~\ref{tab:labels} shows all 111 categories that are exhaustively labeled. The first 20 categories are inherited from PASCAL VOC~\cite{Everingham15}.

\begin{table*}[t]
\adjustbox{width=\linewidth}{
\begin{tabular}{cc|cc|cc|cc|cc|cc}
\toprule
ID & Category & ID & Category & ID & Category & ID & Category & ID & Category & ID & Category\\
\midrule
1 & person & 2 & bird & 3 & cat & 4	& cow & 5 & dog & 6	& horse\\
7 & sheep & 8 & aeroplane & 9 & bicycle & 10 & boat & 11 & bus & 12 & car\\
13 & motorcycle & 14 & train & 15 & bottle & 16 & chair & 17 & dining table & 18 & potted plant\\ 
19 & sofa & 20 & monitor & 21 & stroller & 22 & cabinet & 23 & door & 24 & curtain\\
25 & painting & 26 & shelf & 27 & transformer & 28 & fence & 29 & desk & 30 & bridge\\
31 & lamp & 32 & dome & 33 & railing & 34 & cushion & 35 & box & 36 & column\\
37 & signboard & 38 & tactile paving & 39 & counter & 40 & sink & 41 & barrier & 42 & refrigerator\\
43 & stairs & 44 & case & 45 & crutch & 46 & graffiti & 47 & coffee table & 48 & toilet\\
49 & book & 50 & bench & 51 & barrier gate & 52 & palm & 53 & fruit & 54 & computer\\
55 & arc machine & 56 & parking meter & 57 & light & 58 & truck & 59 & awning & 60 & streetlight\\
61 & booth & 62 & shopping cart & 63 & apparel & 64	& ottoman & 65 & van & 66 & gas bottle\\
67 & fountain & 68 & zebra cross & 69 & toy & 70 & stool & 71 & basket & 72 & bag\\
73 & scooter & 74 & slide & 75 & ball & 76 & food & 77 & tennis court & 78 & pot\\
79 & const vehicles & 80 & sculpture & 81 & vase & 82 & traffic light & 83 & trashcan & 84 & fan\\
85 & plate & 86 & bulletin board & 87 & radiator & 88 & cup & 89 & clock & 90 & flag\\
91 & hot dog & 92 & manhole & 93 & fireplug & 94 & umbrella & 95 & gravestone & 96 & AC\\
97 & mailbox & 98 & push plate & 99 & knife & 100 & phone & 101 & fork & 102 & waiting shed\\
103 & spoon & 104 & faucet & 105 & vending & 106 & frisbee & 107 & banana & 108	& balloon\\
109	& wheelchair & 110 & windmill & 111	& trafficcone & & & & \\
\bottomrule
\end{tabular}}
\caption{Objects from these 111 categories are exhaustively labeled.}
\label{tab:labels}
\end{table*}

\section{Additional Experiment}

In this section, we provide more experiment results, and details about our method.

\subsection{Methods}

\begin{figure}[t!]
\centering
\includegraphics[width=0.9\linewidth]{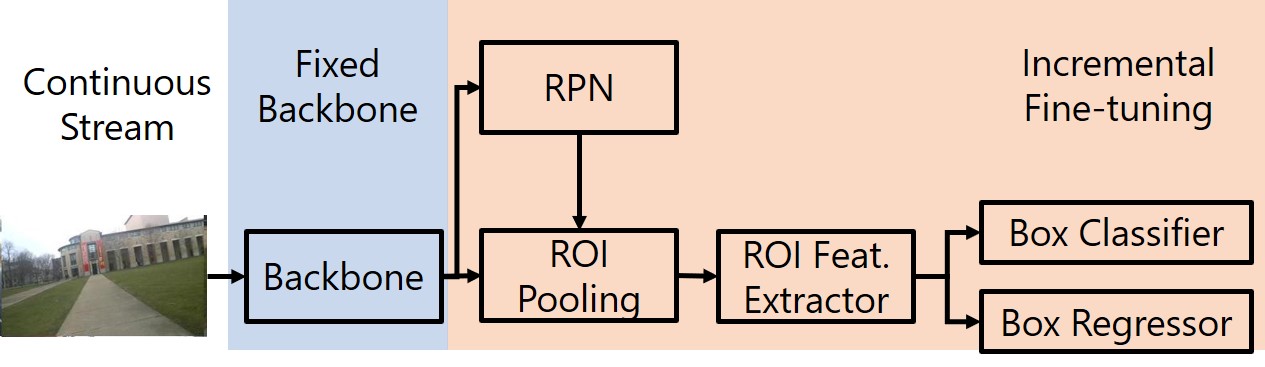}
\caption{Incremental fine-tuning. The entire object detector, including both the feature extractor $\mathcal{F}$ and the box predictor ($\mathcal{R}$ \& $\mathcal{C}$) are pretrained using Pascal VOC. In online continual learning, the feature extractor components are fixed and only the box predictor is fine-tuned.}
\label{fig_supp:baseline_finetune}
\end{figure}

\minisection{Incremental fine-tuning.} The first approach considered is to conduct incremental fine-tuning on the widely used two-stage object detector, Faster R-CNN~\cite{ren2015faster}, which is pretrained on the Pascal VOC dataset. As shown in Figure~\ref{fig_supp:baseline_finetune}, the feature learning components $\mathcal{F}$ include the backbone (e.g., ResNet~\cite{he2016deep}, VGG16~\cite{Simonyan15}), the region proposal network (RPN), as well as a two-layer fully-connected (FC) sub-network as a proposal-level feature extractor. There is also a box predictor composed of a box classifier $\mathcal{C}$ to classify the object categories and a box regressor $\mathcal{R}$ to predict the bounding box coordinates. The backbone features as well as the RPN features are class-agnostic, so we keep them fixed during the incremental fine-tuning process.  Whenever new data comes in, we fine-tune the box classifier and regressor on the new data.

Based on this intuition, we propose to fix the feature extractor and continually fine-tune the box classifier and box regressor. As one can imagine, incremental fine-tuning will lead to fast incremental learning, while suffers from catastrophic forgetting.

\begin{figure}
\centering
\includegraphics[width=\linewidth]{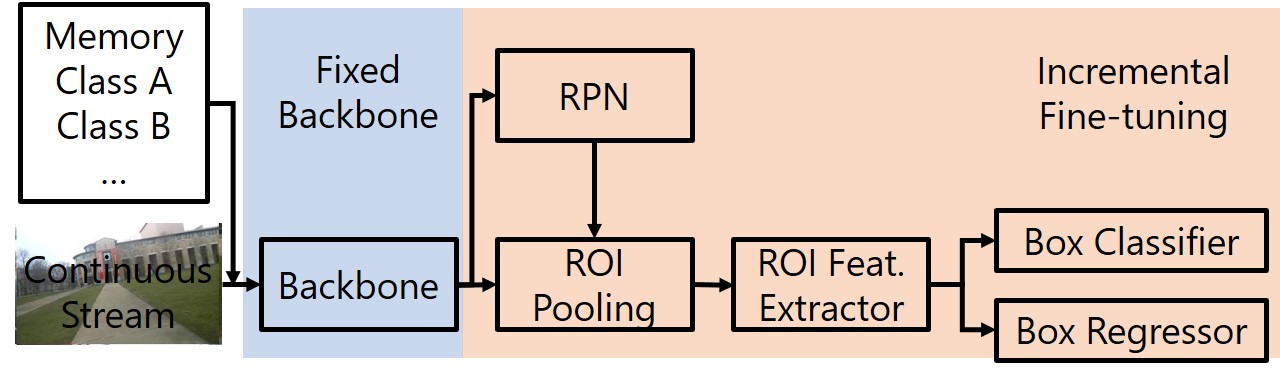}
\caption{Illustraion of iCaRL. During online continual learning, the network parameters are updated by minimizing the loss on both streaming data and memorized prioritized exemplars. During inference, we assign the class label with the most similar prototype.}

\label{fig_supp:baseline_iCaRL}
\end{figure}

\minisection{iCaRL.} To overcome catastrophic forgetting, iCaRL~\cite{rebuffi2017icarl}
introduces three main components in combination: classification by a nearest-mean-of-exemplars rule, selectively storing a subset of the observed examples using feature clustering and representation learning using knowledge distillation and prototype rehearsal (Figure~\ref{fig_supp:baseline_iCaRL}). To save the prioritized exemplar, iCaRL requires a large memory size and tends to fill up the entire memory with streaming data, which does not fit our online continual problem. In contrast, we propose to use a fix memory size (e.g. 10 images) of each category. Whenever iCaRL obtains data from new classes or old classes, it update its feature extractor $\mathcal{F}_\theta$ and the exemplar set $\{\mathcal{M}_k\}$, where k is the index of category. First, iCaRL constructs an augmented training set consisting of the currently available training examples together with the stored exemplars. Next, the current network is evaluated for each example and the resulting network outputs for all previous classes are stored (not for the new classes, since the network has not been trained for these, yet). Finally, the network parameters are updated by minimizing a loss function that for each new image encourages the network to output the correct class indicator for new classes (classification loss), and for old classes, to reproduce the scores stored in the previous step (distillation loss). When the exemplars of each category exceeds the memory budget, we perform \textit{k-means clustering} (where k is the memory budget) and random drop one image from the same cluster.

To predict a label $y$ given a bounding box proposal $x$, it computes a prototype vector for each class observed so far $\{\mu_y\}$, where $\mu_y = \frac{1}{|\mathcal{M}_y|}\sum_{p\in \mathcal{M}_y} \mathcal{F}_\theta(p)$ is the average feature vector of all exemplars for a class $y$. We thus assigns the class label with most similar prototype: $\hat{y} = \mathop{argmin}\limits_{y=1,...,N} ||\mathcal{F}_\theta(x)-\mu_y||$. Compared with incremental fine-tuning, the nearest-mean-of-exemplars rule decouples weight vectors and makes the classification more robust against changes of the feature representation. 

Since iCaRl is designed specifically for image classification, we only train the box classifier using the modules mentioned above. We use incremental fine-tuning to train the box regressor in an class-agnostic manner.

\minisection{GEM.} More directly, GEM~\cite{lopez2017gradient} proposes to minimize the classification loss of current frame subject to not hurting the classification loss of examples stored in the episodic memory through gradient surgery (Figure~\ref{fig_supp:baseline_gem}). Specifically, GEM solves the following problem:

\begin{align}
  minimize_{\theta} & \; \mathcal{L}(\mathcal{F}_\theta(x)-y), \; subject \; to:\\
  \mathcal{L}(\mathcal{F}_\theta(x)-\mathcal{M}_k) & \leq \mathcal{L}(\mathcal{F}^{t-1}_\theta(x)-\mathcal{M}_k) \; for \; all \; k
\end{align}

Mathematically, we rephrase the constraints (2) as:

\begin{align}
  \langle g, g_k \rangle = \langle \frac{\partial\mathcal{L}(\mathcal{F}_\theta(x)-y)}{\partial \theta}, \frac{\mathcal{L}(\mathcal{F}_\theta(x)-\mathcal{M}_k)}{\partial \theta}\rangle \geq 0
\end{align}, for all $k$. However, it is very unlikely that all constrains (3) can be satisfied. Instead, they propose to project the gradient $g$ to the closest gradient $\hat{g}$ satisfying all constrains (3). Therefore, the objectiveness can be written as follows:

\begin{align}
  minimize_{\hat{g}} & \; \frac{1}{2} ||g-\hat{g}||^2, \; subject \; to:\\
  \langle g, \hat{g}_k \rangle & \geq \; for \; all \; k
\end{align}, which can be solved by Quadratic Programming~\cite{frank1956algorithm}. 

Similar as iCaRl, GEM is also designed for image classification. We use incremental fine-tuning to train the box regressor in an class-agnostic manner.

\begin{figure}[t]
\centering
\includegraphics[width=0.9\linewidth]{LaTeX/figures/gem.jpg}
\caption{Illustration of GEM (figure adapted from~\cite{lopez2017gradient}). When learning to predict a new class B, the gradient updates may hurt the performance of the old class A. GEM minimizes the classification loss with constraints
on the gradient updates to not reduce performance of memorized exemplars from the old classes.}
\label{fig_supp:baseline_gem}
\end{figure}

\subsection{Training Details}

For a fair comparison, we use ResNet-50~\cite{he2016deep} pre-trained from ImageNet~\cite{deng2009imagenet} to initialize the backbone of all continual learning algorithms. And the entire model, including feature extractor, box classifier, and box regressor is pretrained on PASCAL VOC~\cite{Everingham15}. During online continual learning, we fine-tuning each method for 10 iterations on each single time stamp. We use SGD~\cite{robbins1951stochastic} for finetuning with a learning rate of $1e-3$ with a weight decay of $1e-6$. To train iCaRL properly, we update the last layer of feature extractor, since iCaRL relies on the feature extractor to make predicions. In contrast, the entire feature extractor is fixed for incremental fine-tuning and GEM. 

\begin{figure}[t]
\centering
\includegraphics[width=0.9\linewidth]{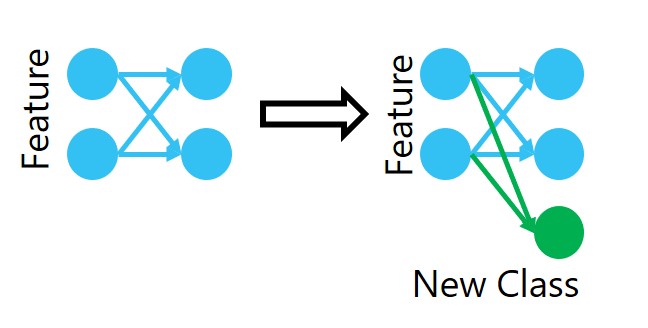}
\caption{When the agent sees instances from previous unseen classes, a new neuron is initialized. The weights of neurons which are used to predict the confidence of old classes are remain unchanged.}
\label{fig_supp:baseline_open_v}
\end{figure}

In the open vocabulary setting, we change the Softmax head~\cite{goodfellow2016deep} of incremental fine-tuning and GEM into multiple Sigmoid head~\cite{goodfellow2016deep}. Thus, the class predictions are decoupled from each other. When the agent sees instances from previous unseen categories, a new neuron is initialized with Kaiming He Initialization~\cite{he2015delving} (Figure~\ref{fig_supp:baseline_open_v}). During prediction, if the confidence of background and all classes are smaller than a threshold, the agent should predict IDK. Experimentally, we set the background threshold to 0.5 and foreground threshold to 0.4.

\subsection{More Results}

\begin{figure*}
\centering
\subfloat[CAPs of Finetuning/iCaRL/GEM/Non-adaptation/Offline training of chair are 12.05/12.73/14.10/0.40/28.33. ]{\includegraphics[width=.9\textwidth]{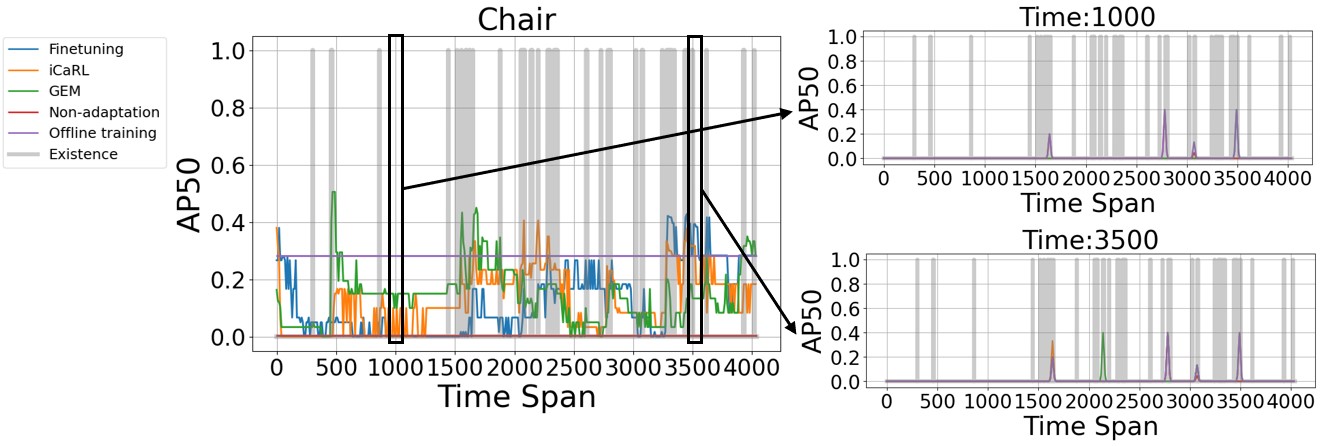}}

\subfloat[CAPs of Finetuning/iCaRL/GEM/Non-adaptation/Offline training of car are 65.18/60.11/57.78/25.17/71.19. ]{\includegraphics[width=.9\textwidth]{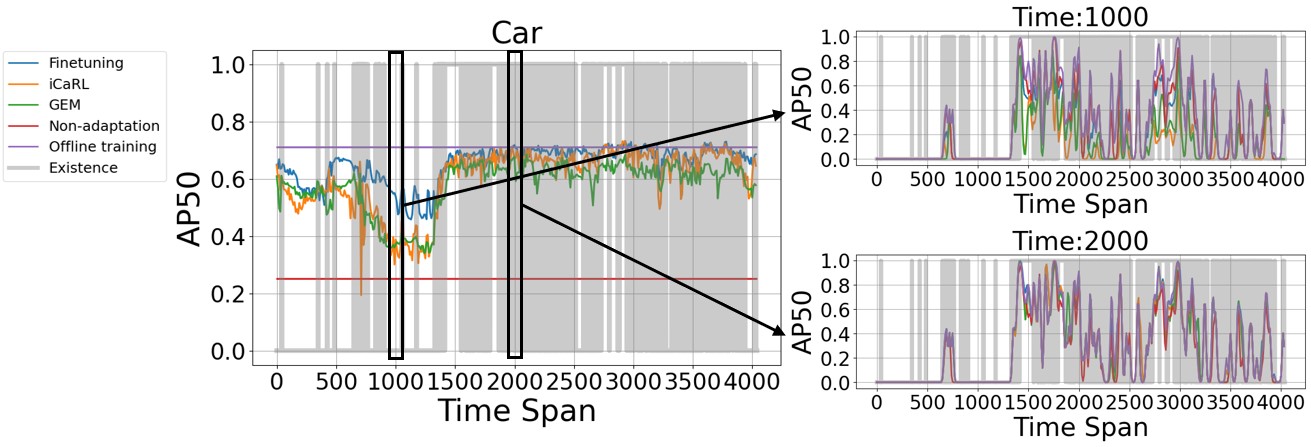}}

\subfloat[CAPs of Finetuning/iCaRL/GEM/Non-adaptation/Offline training of dining table are 36.19/46.05/45.95/2.95/68.39. ]{\includegraphics[width=.9\textwidth]{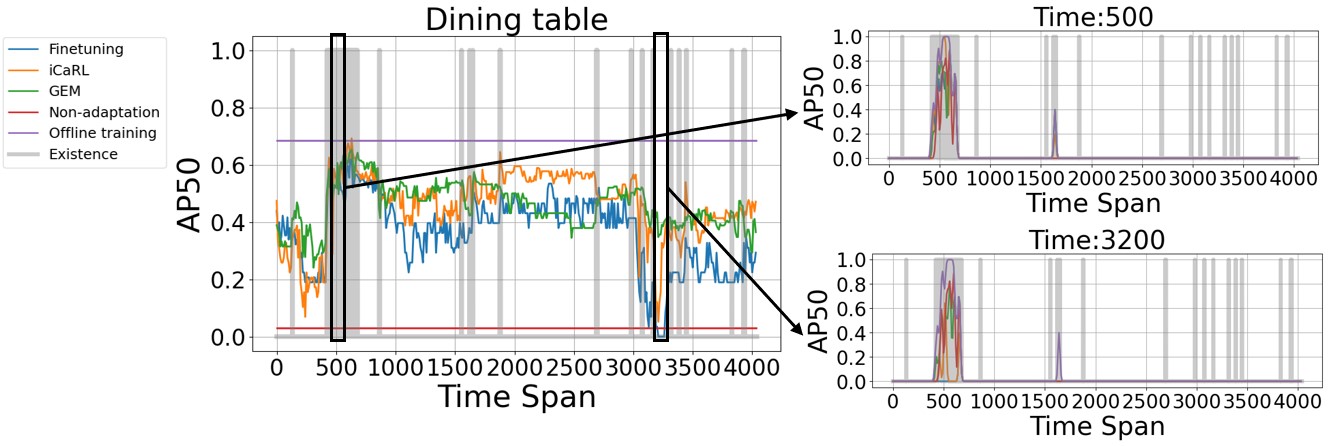}}

\caption{CAP of Apparel on OAK evaluation set, in the order of time stamps.}

\label{fig_supp:result_CAP}
\end{figure*}

\begin{figure*}[t!]
\centering
\centering
\subfloat[Forgetfulness of Finetuning/iCaRL/GEM of bulletin board are 12.07/11.07/5.20. ]{\includegraphics[width=0.31\textwidth]{LaTeX/figures/bulletin board_forgetting.png}}
\hspace{2mm}
\subfloat[Forgetfulness of Finetuning/iCaRL/GEM of case are 4.88/1.58/-0.71.]{\includegraphics[width=0.31\textwidth]{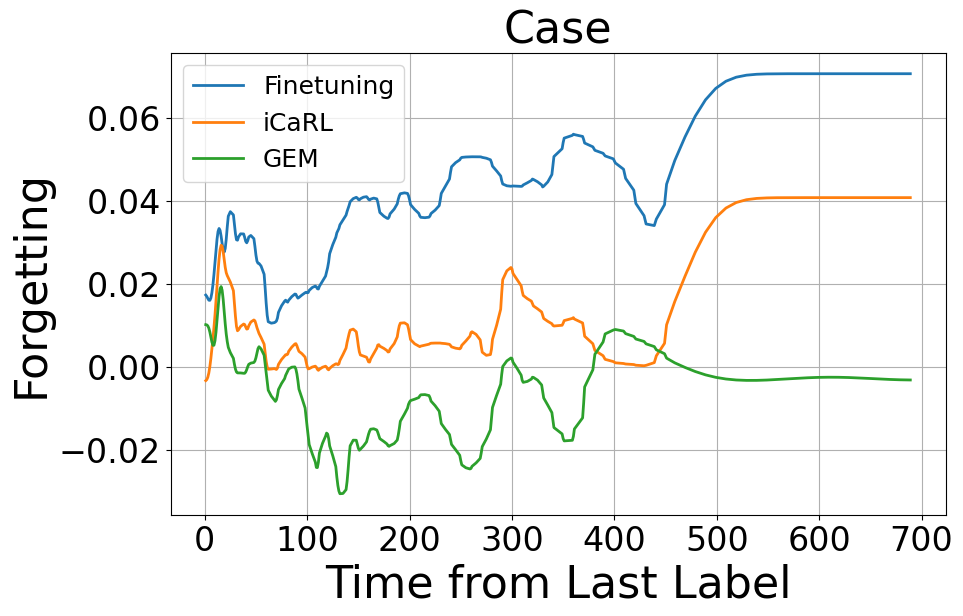}}
\hspace{2mm} 
\subfloat[Forgetfulness of Finetuning/iCaRL/GEM of chair are 8.16/7.26/0.79.]{\includegraphics[width=0.31\textwidth]{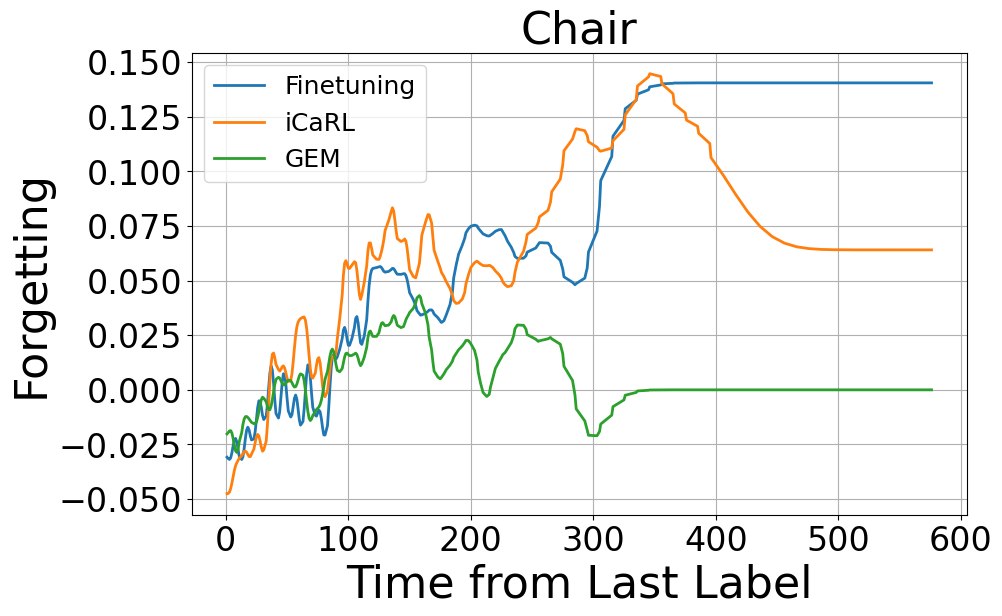}}

\subfloat[Forgetfulness of Finetuning/iCaRL/GEM of food are 31.16/7.50/21.14. ]{\includegraphics[width=0.31\textwidth]{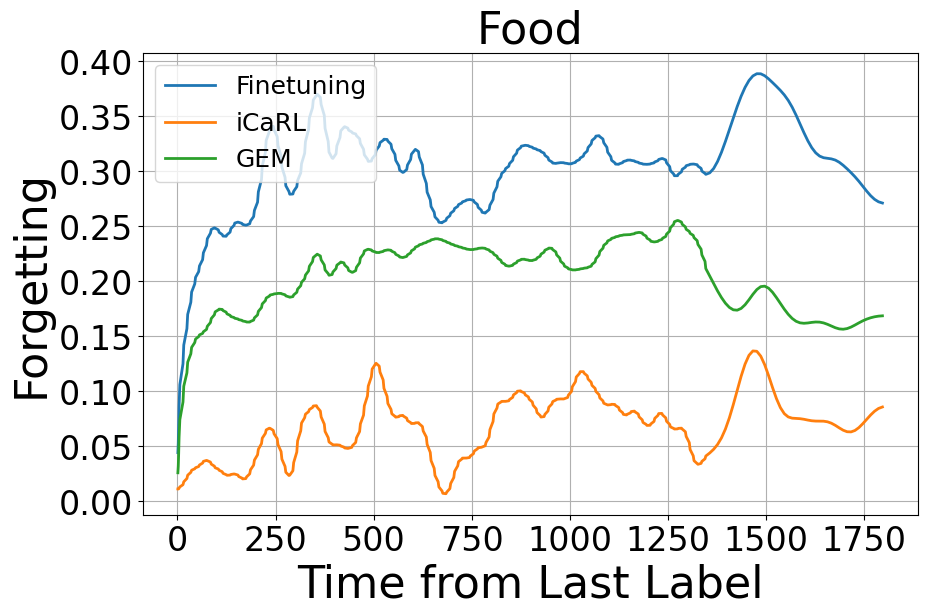}}
\hspace{2mm}
\subfloat[Forgetfulness of Finetuning/iCaRL/GEM of fence are 13.99/10.12/8.97.]{\includegraphics[width=0.31\textwidth]{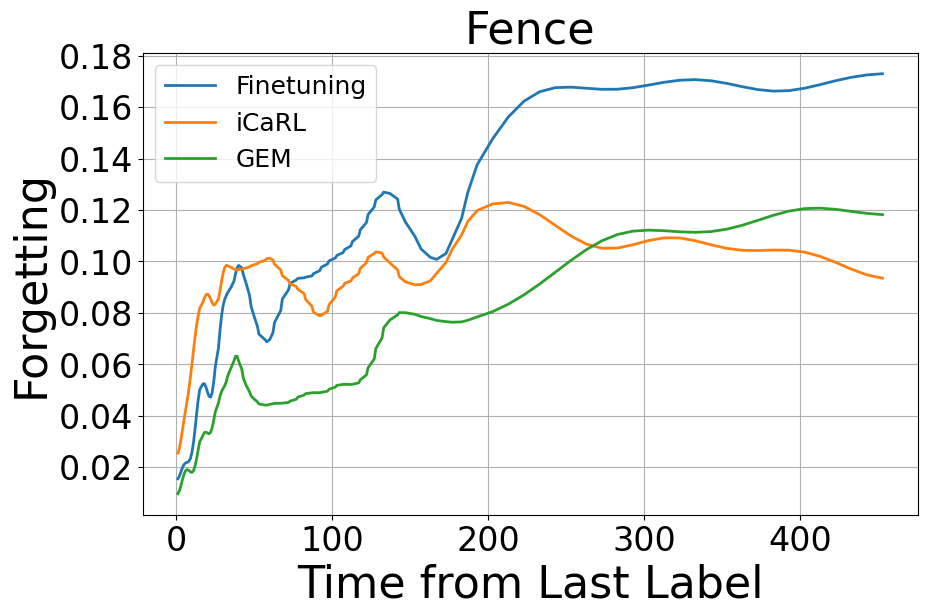}}
\hspace{2mm} 
\subfloat[Forgetfulness of Finetuning/iCaRL/GEM of signboard are 10.38/10.41/3.59.]{\includegraphics[width=0.31\textwidth]{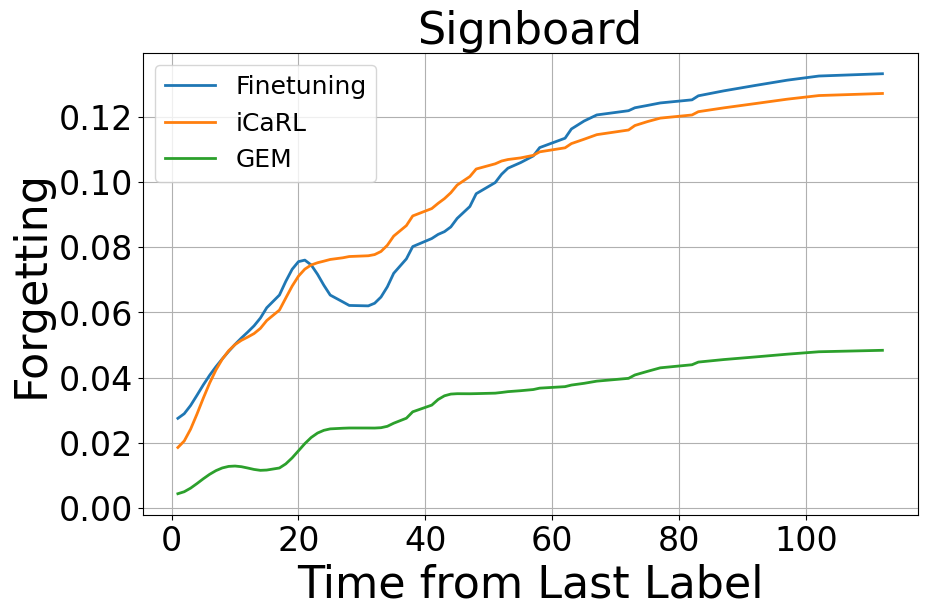}}
\caption{Forgetfulness of sample categories. We provide the evaluation curve on the test set under different time stamps.}
\label{fig_supp:result_F}
\end{figure*}

In this section, we show the CAP (Figure~\ref{fig_supp:result_CAP}) and Fogetfulness (Figure~\ref{fig_supp:result_F}) of more categories on OAK. 

\section{Visualization}

Please see the attached video for more visualization.

% 5-min
% What we did
% show case: using label -> represent the same idea in figure 1, other settings are aritifical.
% show detection performance -> moving camera / blurring / few shot / forgetting / domain shift

\section{Code}

Please refer to the attached code to see our implementation.

{\small
\bibliographystyle{ieee_fullname}
\bibliography{references}
}